\documentclass{article}
\usepackage{spconf,amsmath,graphicx}
\usepackage{comment}

\usepackage{booktabs}
\usepackage{hyperref}

\usepackage{xcolor}

\title{Efficient Transformer for Direct Speech Translation}
\name{Belen Alastruey, Gerard I. Gállego, Marta R. Costa-jussà\thanks{This work is supported by the project ADAVOICE PID2019-107579RB-I00 / AEI / 10.13039/501100011033 and by the European Research Council (ERC) under the European Union’s Horizon 2020 research and innovation programme (grant agreement No. 947657).}}
\address{TALP Research Center, Universitat Politècnica de Catalunya, Barcelona}
\begin{document}
%
\maketitle
\begin{abstract}
The advent of Transformer-based models has surpassed the barriers of text. When working with speech, we must face a problem: the sequence length of an audio input is not suitable for the Transformer. To bypass this problem, a usual approach is adding strided convolutional layers, to reduce the sequence length before using the Transformer. 

In this paper, we propose a new approach for direct Speech Translation, where thanks to an efficient Transformer we can work with a spectrogram without having to use convolutional layers before the Transformer. This allows the encoder to learn directly from the spectrogram and no information is lost.
We have created an encoder-decoder model, where the encoder is an efficient Transformer –the Longformer–  and the decoder is a traditional Transformer decoder. 
Our results, which are close to the ones obtained with the standard approach, show that this is a promising research direction.
\end{abstract}
\begin{keywords}
Direct Speech Translation, Efficient Transformer, Longformer
\end{keywords}

\section{Introduction}
\label{sec:intro}

    The task consisting of translating speech into a written form in another language is popularly known as Speech Translation (ST). The first model of this type consisted in the concatenation of two independent models, forming what nowadays is known as cascade system \cite{first-cascade-st}. The first module, an Automatic Speech Recognition model (ASR), writes a transcription of the spoken sentence, and the second one, a Machine Translation (MT) model, translates the transcription to another language. But in the last few years, new models based on end-to-end architectures have emerged. These models are capable of translating from audio to text, without the need to go through the intermediate step of transcription. These models, also known as direct ST systems, have rapidly evolved and, nowadays, they can reach the same state-of-the-art results than cascade models \cite{IWSLT2020, cascade_vs_end2end}. Nevertheless, results provided by both cascade and end-to-end architectures are far from optimal, and therefore these research fields are still under development.
    
    In the recent years, Transformer-based language models have gained popularity and have revolutionized Natural Language Processing (NLP) \cite{transformer, bert, gpt3}. However, the Transformer is suitable for short sentences, but not for long texts, because of its quadratic complexity. To overcome this problem, many efficient Transformer variants have been proposed, with linear complexity instead of quadratic \cite{efficient-transformers,long-range-arena}. Moreover, in the last few years, the Transformer has surpassed the barriers of NLP, and it has also been applied in other fields, such as computer vision \cite{transformer-computervision1,transformer-computervision2} and speech processing \cite{first-asr-transformer,s-transformer,transformer-speech}. In Speech-to-Text context, a standard approach is working with previously extracted audio features, like the mel-spectrogram. The sequence length of these representations is approximately an order of magnitude longer than usual text sequence lengths. Hence, working with such large speech sequences with the original Transformer has a dramatic impact to its computational complexity. To overcome this problem, a common approach is to use convolutional layers with stride before the Transformer encoder, to reduce the sequence length of the inputs \cite{first-asr-transformer,s-transformer}.
    
    The objective of this work is contributing to ST research, proving the feasibility of using efficient Transformers for audio inputs, without the need of adding convolutional layers. The motivation of using an efficient Transformer is that they were designed to process long text sequences, which can be extrapolated to speech mel-spectrograms. In particular, we use the Longformer \cite{longformer}, with a sliding window self-attention that we believe it could be profitable for audio processing. Our goal is to take advantage of the lower complexity of this model and create a Speech-to-Text Transformer where the Longformer deals with the audio input. We believe the training could benefit from this approach, since it lets the model learn directly from the spectrogram and no information is lost in the convolutional layers. On the negative side, a possible complication derived of this system is that cross-attention between the encoder and the decoder could be hindered by a mismatching between the input and the output sequence lengths.

\section{Related Work}
\label{sec:relatedwork}

    The first models for direct ST \cite{first-end2end-st1,first-end2end-st2} consisted of encoder-decoder architectures made up of recurrent neural networks, like LSTMs \cite{lstm}, inspired by LAS \cite{las}. The introduction of the Transformer for MT \cite{transformer} also influenced the ST field. With the good results shown in NLP \cite{bert} it was natural to consider using this architecture also for speech tasks \cite{first-asr-transformer}, including ST \cite{first-end2end-st-transformer}. However, the quadratic complexity of the Transformer's attention computation makes it especially difficult for this architecture to process speech inputs. These kinds of sequences are about an order of magnitude longer than text inputs, therefore, the computational cost of training the model can rise critically. Hence, a common approach in ST systems is to add convolutional layers before the Transformer encoder that reduce the input sequence length \cite{s2t_transformer}. Other systems also include 2D self-attention layers and a distance penalty in the attention, to bias it towards the local context \cite{s-transformer}.

    The computational complexity of the Transformer's attention matrix is $O(n^2)$, where $n$ is the sequence length. This quadratic cost not only hinders working with speech inputs, but also with long text sequences, such as documents. Recently, many efforts have been made to overcome this issue, proposing new architectures known as efficient Transformers \cite{efficient-transformers,long-range-arena}. The Longformer \cite{longformer} and the Big Bird \cite{bigbird} models modify the attention matrix with patterns such as sliding, global and random attentions. Other models, like the Reformer \cite{reformer} and the Routing Transformer \cite{routing_transformer}, only compute attention weights for those queries and keys which are more related. The authors of the Linformer \cite{linformer} state that the attention matrix is low-rank, so they projected the keys and the values to reduce the size of the attention matrix. The Synthesizer \cite{synthesizer} directly avoids computing token-token interactions by learning synthetic attention weights.

    In the speech processing field, concretely in ASR, some studies have explored modifications of the attention mechanism, with different motivations. Some researchers applied a self-attention layer with augmented memory, to use information beyond the whole utterance level \cite{transformer-augmented-memory-asr}. Others focused in reducing the number of parameters of the Transformer by modifying the way the queries and keys of the self-attention are computed \cite{transformer-mod-asr}. The work which is more related to ours \cite{synthesizer-asr} used a Synthesizer model \cite{synthesizer} for ASR, improving the performance and reducing its complexity. However, to the best of our knowledge, our work is the first using an efficient Transformer for direct ST.

\section{Longformer}
\label{sec:longformer}

\begin{figure}[h]
        \centering
        \includegraphics[width=.7\linewidth]{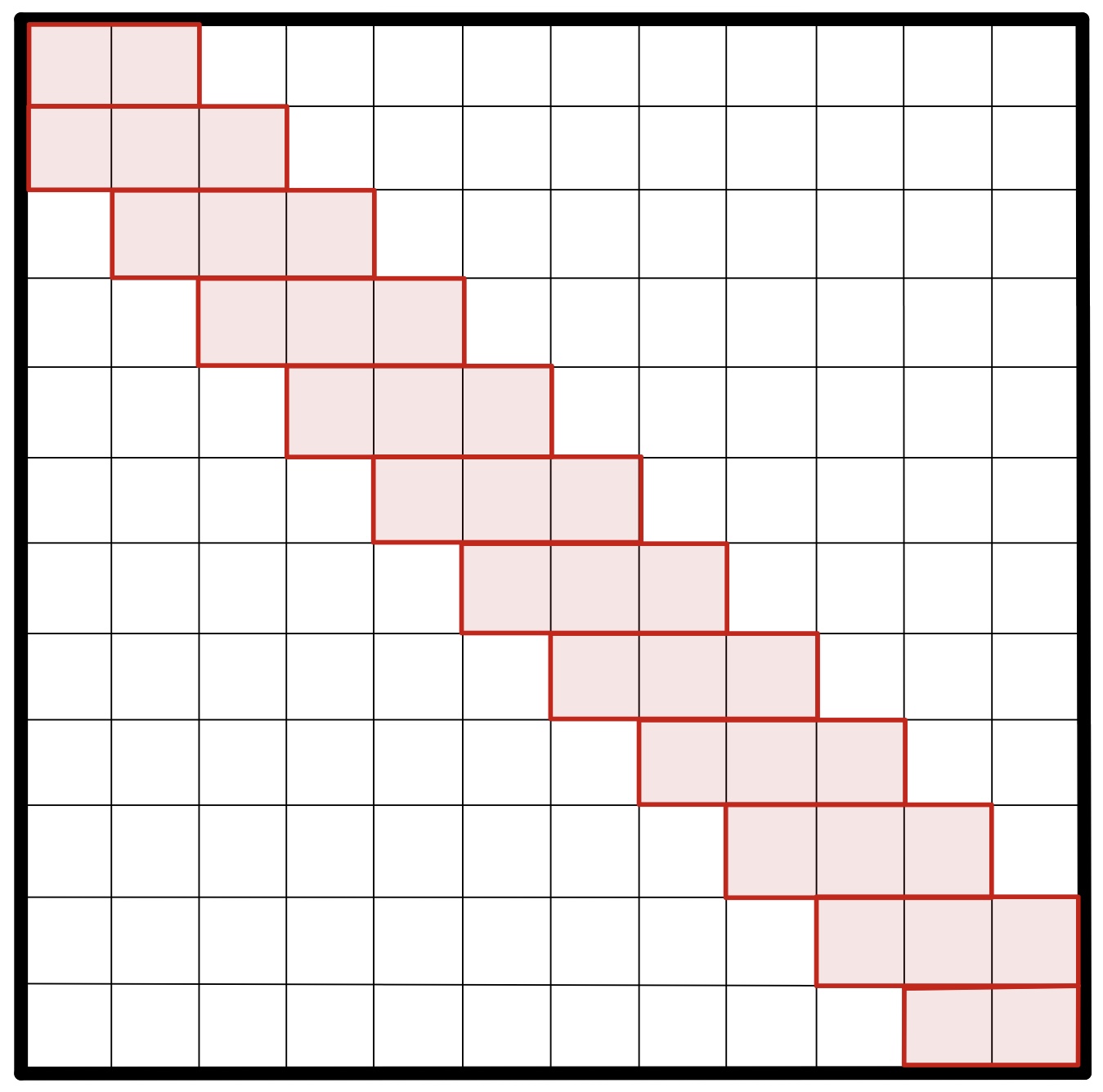}
        \caption{Sliding window pattern, used by the Longformer self-attention.}
        \label{fig:window}
    \end{figure}

    In this work, we use the Longformer \cite{longformer}, a variation on the original Transformer, which achieves a reduction in the complexity of the attention computation, from quadratic to linear. To achieve this improvement, the Longformer defines a pattern in the attention matrix, specifying, for each token combination, the attention weights that need to be computed. Removing some attentions between tokens, reduces the number of operations; hence, the algorithm scales linearly with the input sequence length. Longformer's attention pattern consists of the following components:
    
    \begin{itemize}
        \item \textbf{Sliding Window}: It is the main component of the attention pattern, and it relies on the importance of local context. With this component, an attention window of fixed size is placed around each token (Figure \ref{fig:window}). 
        Given a fixed window size $w$ each token attends to the $\frac{1}{2}w$ tokens on each of its sides. In addition to the local context, adding several stacked attention layers achieves a similar effect as CNNs, that allows the last layers to receive information from a large region of the input and not only from the tokens inside the window. Stacking $l$ attention layers provides a receptive field of size $l \times w$ at the top layer.
        \item \textbf{Dilated Sliding Window}: It is a variant of the sliding window, which allows to increase each token's attention range without increasing the complexity. For every token, the dilated sliding window also attends to $\frac{1}{2}w$ tokens on each side but leaves gaps of size $d$. Consequently, the receptive field in the last layer has a size of $l \times d \times w$.
        \item \textbf{Global Attention}: In some NLP tasks, there are special tokens, such as [CLS] in the case of BERT, that would not be attended enough just by using the sliding window. In these cases, global attention is added to pre-selected input tokens. This global attention is symmetric: the selected token attends to every other token and vice versa.
    \end{itemize}
    
\section{Speech-to-Text Longformer}

    Although translating long texts is not the objective of this work, we believe that efficient Transformers could be useful to deal with spoken sentences, since they can help to address the issues caused by long sequence lengths. We have considered different efficient Transformer models such as the Big Bird \cite{bigbird}, the Linformer \cite{linformer}, the Longformer \cite{longformer}, the Reformer \cite{reformer}, the Routing Transformer \cite{routing_transformer} and the Synthesizer \cite{synthesizer} (\S\ref{sec:relatedwork}). Finally, we chose the Longformer because of its attention pattern (\S\ref{sec:longformer}). We believe that the sliding window approach may be profitable for audio processing. We also chose it because we needed an encoder-based model, since we just wanted to substitute the encoder from the original Transformer architecture.
    
\begin{figure}[t]
        \centering
        \includegraphics[scale=0.7]{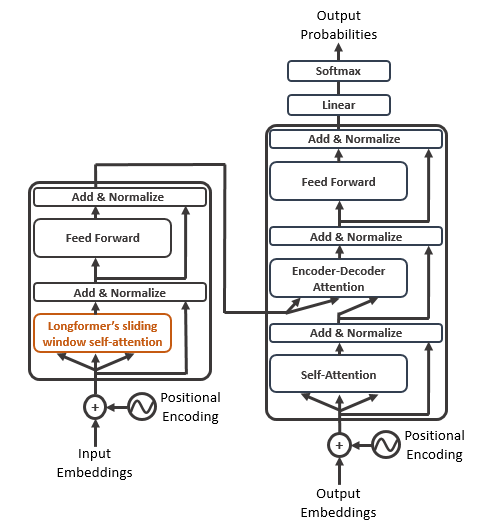}
        \caption{System overview. We replace the Transformer encoder self-attention by the Longformer's.}
        \label{fig:our_model}
    \end{figure}

    Our system, the Speech-to-Text Longformer, is composed of an encoder consisting of a Longformer model, and a base Transformer decoder (Figure \ref{fig:our_model}). We use the same number of layers in the encoder regarding the baseline Transformer model (\S\ref{sec:experiments}) for a proper comparison. We use a regular sliding window as the self-attention pattern, since the dilated version is not available in the implementation of the Longformer we used. We neither used global attention, since it is not suitable for the tasks we designed the system for. We explored different sizes for the sliding window self-attention.
    
    However, we hypothesized that our model could have an inconvenience. In ST, as in every translation task, when computing the encoder-decoder attention, the input and the output sentence are aligned. But in this case, while the input is a spectrogram, the output is text, and this sequence length mismatch can make the alignment too complicated. This is not a problem in other systems, since they reduce the sequence length before the Transformer and, therefore, the sequence length mismatch is much smaller, which eases the task of alignment.
    
    To solve this problem, we build an alternative system, adding a convolutional layer with stride after the Transformer encoder. This reduces the encoder output length by half, making it more similar in size to the output text sentence, eventually solving the length discrepancy issue. This approach is close to the one used by other systems using convolutions, but in our case it is placed after the encoder, instead of before. We believe this could be profitable, since we let the encoder learn directly from the spectrogram, without previous information loss. At the same time, we reduce the complexity of the model and address the alignment problem with the sequence lengths mismatch.

\section{Experiments}
\label{sec:experiments}

    In this section, we describe the different experiments that we carried out, including the dataset, the experimental settings, and the analysis of the results.
    
    We compare our systems with the Speech-to-Text Transformer model available in Fairseq \cite{s2t_transformer}, to evaluate the performance of our systems with respect to a baseline. In particular, we use the small architecture, which is the one with reported results\footnote{Fairseq Speech-to-Text example of ST: \url{https://github.com/pytorch/fairseq/blob/master/examples/speech_to_text/docs/mustc_example.md}}.

    \subsection{Data}

        We train our system with the MuST-C dataset \cite{mustc}. This dataset was created after the popularity growth of End-to-end ST, with the aim to confront the scarcity of public data that researchers were facing when training these new type of models. Cascade solutions have many and varied data to train each of their modules, but End-to-end ST datasets were small and of limited language coverage. MuST-C dataset is a Multilingual ST Corpus (MuST-C) built using TED talks in English. The dataset includes a corpus for translation from English into 14 different languages, that belong to different families, including Dutch, French, German, Italian, Portuguese, Romanian, Russian, or Spanish. It contains at least 237 hours of transcribed recordings (430 on average) for each of the available languages. Furthermore, the data is free and of good quality, and includes a variety of topics and speakers. 
        
        For our work, we use the English-German split, including 408 hours of English speech and 234k English-German paired sentences.

    \subsection{Settings}

        Before discussing the conducted experiments and their results, let us describe some decisive conditions, such as the details of the model implementation and the training parameters.

        \subsubsection{Speech-to-Text Longformer Implementation}
            To build our model, we use the encoder of the Longformer model available in Hugging Face \footnote{\url{https://huggingface.co/transformers/model_doc/longformer.html}}, and a regular Transformer decoder available in Fairseq \cite{s2t_transformer}.
            In order to get the most realistic comparison possible between our results and the ones obtained with the Speech-to-Text Transformer \cite{s2t_transformer}, we try to create a model as similar as possible to theirs.
            We build our model with 12 encoder layers and 6 decoder layers. We apply a normalization layer before each decoder layer. In both the encoder and the decoder, we use 4 attention heads, an embedding dimension of 256, 2048 dimensions in the FFN layers, and sinusoidal positional encodings. The decoder output dimension is 256, the same as the decoder embedding dimension. We use a dropout probability of 0.1 in both the attention weights and in the FFN activations. We use ReLU as the activation function for the FFNs.
            
            Additionally, there are some extra parameters, that are specific of our model, regarding the size of the attention window and the convolutional layer we apply to reduce sequence length. The latter consists a 1D convolutional layer, with a kernel of size 5, a stride of 2, and with the same number of output channels than input channels. On the other hand, the attention window size is defined specifically for each experiment.

        \subsubsection{Training parameters}
            To ensure a reliable comparison, we perform all ASR and ST experiments under the same conditions and parameters. Specifically, we try to use the same parameters as in the implementation by \cite{s2t_transformer}, when possible.
            In ASR trainings we use 4 CPUs and 2 workers to load the data. We fixed a maximum of 20000 tokens per batch. We used Adam optimizer and a learning rate of $1 \cdot 10^{-3}$ with an inverse square root scheduler. We applied a warm-up for the first 10000 updates.  We clipped the gradient to 10 to avoid exploding gradients. We used label smoothed Cross-entropy as a loss function, with a smoothing factor of 0.1. We used an update frequency of 16, simulating the use of 16 GPUs. We fix a maximum of 100000 updates for every training.
            In ST trainings we use the same parameters as for ASR, but for the learning rate, that is $2 \cdot 10^{-3}$, as done in \cite{s2t_transformer}.
            We conducted the training of all our experiments in an NVIDIA GeForce RTX 2080 Ti GPU.

        \subsection{Experiments description}

            Apart from comparing our system to the Fairseq's Speech-to-Text Transformer, we also want to study the influence of the variables of our model. For this reason, we train different combinations of attention window sizes and the application or not of the convolutional layer after the encoder.
            
            Before training our models for ST, we perform a pre-training step of the encoder. This process consists of training the whole system for ASR, and then using just the pre-trained encoder in the following ST training. This is a common technique when building these kinds of systems \cite{asrpretraining-st}. The motivation of this method is that, in ST, the encoder has to learn two very different tasks at the same time: the acoustic and the semantic modeling. Meanwhile, in ASR the encoder can focus on learning the acoustic modeling, since alignment is monotonic and semantics are not the main issue. Moreover, the amount of data available for ASR training is higher than for ST, what can improve the results.

        \subsubsection{Automatic Speech Recognition}
            For the ASR task we experimented with and without the convolutional layer, and with multiple window sizes (512, 76, 60, 48). We train the models using English-German split of MuST-C dataset. In table \ref{tab:asr} we find the best WER of each experiment. Some of the experiments, such as the ones with a window size of 512, presented instability and convergence problems, and therefore, do not appear in the table. 
            
            \begin{table}[h]
                \centering
                \begin{tabular}{lcc}
                    \toprule
                    Model                               & Window        & WER ($\downarrow$) \\
                    \midrule
                    s2t\_transformer           & -             & 13.31             \\
                    s2t\_longformer $\dagger$  & 76            & 15.00             \\ 
                    s2t\_longformer            & 60            & 14.99             \\
                    s2t\_longformer $\dagger$  & 48            & 14.72             \\
                    s2t\_longformer           & 48            & 15.12             \\
                    \bottomrule
                \end{tabular}
                \caption{ASR results on the MuST-C test subset of the English-German split. $\dagger$: With the convolutional layer.}
                \label{tab:asr}
            \end{table}
            

        \subsubsection{Speech Translation}
            In this section, we describe the ST experiments that have been carried out. In order to train a model for ST we needed the pre-trained encoder obtained after the ASR training. Furthermore, it is also worth noting that since ST is a harder task than ASR (because of alignment), a model that has not worked well for ASR is highly unlikely to work for ST.
            For these reasons, we experimented again with and without the convolutional layer, and with multiple window sizes (76, 60, 48), but only with the models that worked for ASR.
            We train the models using English-German split of MuST-C dataset. In table \ref{tab:st} we find the best BLEU of each experiment.

            \begin{table}[h]
                \centering
                \begin{tabular}{lcc}
                    \toprule
                    Model                               & Window        & BLEU ($\uparrow$) \\
                    \midrule
                    s2t\_transformer           & -             & 22.41             \\
                    s2t\_longformer $\dagger$  & 76            & 20.64             \\ 
                    s2t\_longformer           & 60            & 20.34             \\
                    s2t\_longformer $\dagger$  & 48            & 20.45             \\
                    s2t\_longformer            & 48            & 20.49             \\
                    \bottomrule
                \end{tabular}
                \caption{ST results on the MuST-C test subset of the English-German split. $\dagger$: With the convolutional layer.}
                \label{tab:st}
            \end{table}
            
            \begin{figure}[h]
                \centering
                \includegraphics[width=\linewidth]{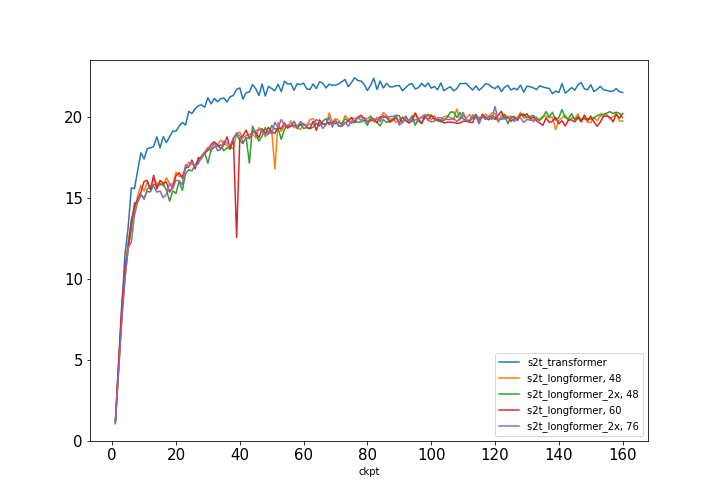}
                \caption{ST results over epochs (BLEU).}
                \label{fig:ST_graph}
            \end{figure}

    \subsection{Results Analysis}

        From the ASR training, we see that using a large attention window can harm dramatically the system training. We hypothesize that using such long windows do not let the encoder focus properly on the audio feature extraction. We can see this, especially, when using the window of size 512, the default value in the Hugging Face implementation of the Longformer. A possible explanation for this is that this model has been created for text (long documents), not for speech, and therefore the optimal parameter is not necessarily the same for both tasks.  We believe that this attention window is too big for a speech task, since the attention is applied to an interval of approximately 5 seconds, which could be too wide for performing the audio features extraction. Therefore, we believe that narrower windows, such as 76, 60 or 48 are more suitable for audio processing. Nevertheless, we believe that this window size could be too small to work well in an ST task, since it does not allow the model to learn meaningful relations between distant words. This is not very relevant in ASR tasks, because alignment is monotonic, but can be a problem in future ST trainings where alignment with the rest of the words in the sentence is fundamental. Additionally, we realize that the Longformer instability can be an obstacle for the training process, especially aggravated with long window sizes, that make the model more complex. Even so, we don't observe much difference in the performance of our models, despite the differences in the window size and the application or not of the convolutional layer. The results of these models are between 1.41 and 1.81 points behind the baseline (Table \ref{tab:asr}).
        
        We use the pre-trained ASR encoders of these four models to ease the convergence of the ST training. Hence, these models do not suffer from instability during the ST training. As in the case of ASR, our models do not differ much between them, but they do not reach the baseline results (Figure \ref{fig:ST_graph}). In this case, they fall between 1.77 and 2.07 points behind the baseline (Table \ref{tab:st}). Finally, it is remarkable that, in contrast to what we hypothesized, adding a convolutional layer is not especially useful in neither ASR nor ST. So, unexpectedly, the alignment between speech and text has not been a problem for the performance of the model. Instead, we had to face other issues that we did not expect: large attention window sizes caused training instability.

\section{Conclusions}

    Direct ST is a research field in development, since current results are still far from optimal. Therefore, there is still work to be done before these algorithms can be used in real-life applications.
    
    This paper shows a variation of the original Transformer that makes it suitable for ST tasks. It replaces the encoder's self-attention with the attention pattern proposed by the Longformer, based on a sliding window. From our point of view, this pattern can be useful for audio processing. This allows the model to work directly with a spectrogram, without losing any information in the convolutional layers before the Transformer, that other systems implement \cite{s-transformer,s2t_transformer}.
    
    Our model did not reach the baseline results, but got a close performance: a WER of $14.72$ (compared to $13.31$ from the baseline system) and a BLEU score of $20.64$ (compared to $22.41$ from the baseline system), which we consider a great starting point for a promising research path.
    
    After these results, we believe it would be appropriate to try a different approach when defining the attention window size, trying the dilated sliding window or a window of variable length. Additionally, another future work could be using Big Bird's attention pattern \cite{bigbird}, so we could study the effect of adding random attention to the Longformer approach. Finally, we could try other efficient Transformers, such as the Linformer, that has been tried in encoders and therefore could be suitable for a model like ours.

\bibliographystyle{IEEEbib}
\bibliography{refs}

\end{document}